\pdfoutput=1

\documentclass[11pt]{article}

\usepackage[final]{acl}

\usepackage{times}
\usepackage{latexsym}

\usepackage[T1]{fontenc}

\usepackage[utf8]{inputenc}
\usepackage{listings}
\definecolor{backcolour}{rgb}{0.95,0.95,0.95}
\definecolor{codegray}{rgb}{0.5,0.5,0.5}
\lstdefinestyle{mystyle}{
    backgroundcolor=\color{backcolour},
    basicstyle=\ttfamily\footnotesize\color{codegray},
    breakatwhitespace=false,         
    breaklines=true,                 
    captionpos=b,                    
    keepspaces=true,                 
    tabsize=4
}
\usepackage{microtype}

\usepackage{inconsolata}

\usepackage{graphicx}

\usepackage{algorithm}
\usepackage{algorithmic}
\usepackage{graphicx}

\newcommand*{\img}[1]{%
    \raisebox{-.2\baselineskip}{%
        \includegraphics[
        height=1.1em,
        width=1.1em,
        keepaspectratio,
        ]{#1}%
    }%
}
\usepackage{booktabs}
\usepackage{hyperref}

\usepackage{textcomp}
\usepackage{subcaption}

\usepackage{newfloat}
\usepackage{listings}
\floatstyle{ruled}
\newfloat{listing}{tb}{lst}{}
\floatname{listing}{Listing}
\title{OneLove beyond the field - A few-shot pipeline for topic and sentiment analysis during the FIFA World Cup in Qatar}

\author{
    Christoph Rauchegger\textsuperscript{\rm 1} \and Sonja Mei Wang\textsuperscript{\rm 2} \and Pieter Delobelle\textsuperscript{\rm 3} \\
    \textsuperscript{\rm 1} Technische Universit\"at Berlin, Germany \\
    \textsuperscript{\rm 2} University of Wuppertal, Germany\\ 
    \textsuperscript{\rm 3} Department of Computer Science, KU Leuven; Leuven.AI, Belgium\\
}

\usepackage{bibentry}

\begin{document}

\maketitle

\begin{abstract}
The FIFA World Cup in Qatar was discussed extensively in the news and on social media. Due to news reports with allegations of human rights violations, there were calls to boycott it. Wearing a OneLove armband was part of a planned protest activity. Controversy around the armband arose when FIFA threatened to sanction captains who wear it. To understand what topics Twitter users Tweeted about and what the opinion of German Twitter users was towards the OneLove armband, we performed an analysis of German Tweets published during the World Cup using in-context learning with LLMs. We validated the labels on human annotations. We found that Twitter users initially discussed the armband's impact, LGBT rights, and politics; after the ban, the conversation shifted towards politics in sports in general, accompanied by a subtle shift in sentiment towards neutrality. Our evaluation serves as a framework for future research to explore the impact of sports activism and evolving public sentiment. This is especially useful in settings where labeling datasets for specific opinions is unfeasible, such as when events are unfolding.
\end{abstract}

\maketitle

\section{Introduction} 

In December 2010, it was announced that Qatar will host the FIFA World Cup in 2022 \cite{qatar_announce_bbc}. This announcement was met with concerns and criticism, including the effect of high temperatures during June and July \cite{matzarakis2015sport}, and accusations of bribery \cite{ny_times}. Important political topics were allegations of human rights violations, such as withholding wages from migrant workers and their unexplained deaths \cite{heerdt2023lessons,hr_qatar_post}, as well as abuse endured by LGBT people \cite{dw_abuse_lgbt, hr_qatar_lgbt}. 

As one of the biggest sport events of 2022, the FIFA World Cup in Qatar was followed by football fans and social media users who expressed their enthusiasm, but also critique towards this sporting event on the microblogging platform X, formerly known as Twitter \cite{pak2010Twitter,fan2020Twitter}. FIFA claims that there were over five billion ``engagements" with the World Cup on social media \cite{fifa_engagement}. 
The FIFA World Cup is one of the most followed sporting events worldwide -- a mega-event \cite{muller2023peak}. 

\begin{figure*}
    \centering
    \includegraphics[width=0.92\linewidth]{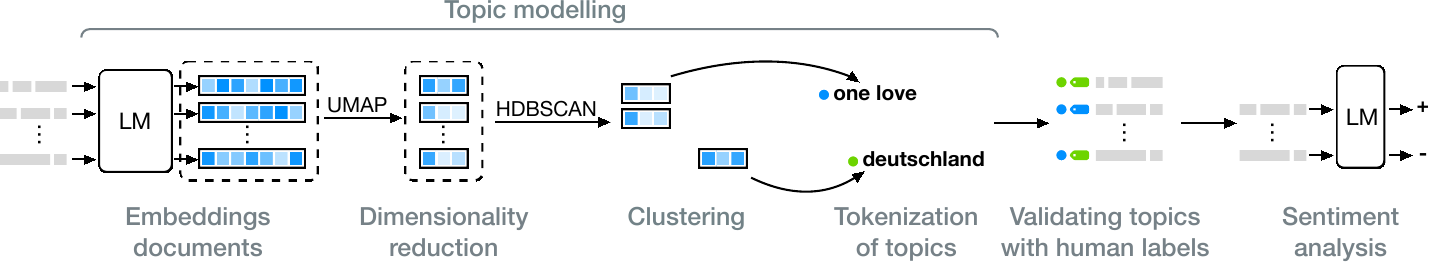}
    \caption{Topic modeling and sentiment analysis pipeline.}
    \label{fig:pipeline}
\end{figure*}

While calls for boycotts became public before the World Cup and initiatives like ``Boycott Qatar 2022'' were established \cite{sportschau_boycott}, there were planned protest activities during the World Cup, such as the OneLove captain's armband. Ten European teams announced their intention to wear the colored armband \cite{one_love_before}. Announcements of sanctions made national teams cancel these plans, with threats of receiving a yellow card for wearing it becoming known on November 21st, 2022 \cite{dw_onelove}. Teams released statements expressing that it had been their intention to wear the armband to stand for inclusion, diversity and mutual respect, but chose not to in order to avoid negatively impacting the players \cite{cnn_onelove}. Instead of wearing the armband, the German national team held their hands in front of their faces to make a statement \cite{zdf_onelove_hand}. The sanctions against the German team's protest activity and the backing down of teams were discussed on social media, such as Twitter.

We conducted an analysis of 132k Tweets about the FIFA World Cup in Qatar to analyze topics dominating discussions on this event and opinions on one particular topic, namely the OneLove armband. By conducting this analysis, we also evaluate if our proposed method can replace supervised pipelines to analyze and measure public opinions.
Supervised machine learning models play a crucial role in measuring sentiment online~\citep{dang2020sentiment} by classifying user-generated texts 
\citep{lheureux2023predictive, rustam2021performance, scott2021measuring}. However, training these models requires labeled data, which involves significant time and resources. Thus, some projects are limited to using existing models, for instance when there is no annotation budget, or when monitoring events or crises, where there is no time to create a labeled dataset~\citep{bruyne2024crisis}. Another limitation is that most studies of sentiment of specific events on social media happen after the event has passed, instead of while an event is unfolding, which is possible with our proposed method. 

We specifically aim to answer the following research questions, both related to the World Cup in Qatar and for monitoring opinions on social media:
\vspace{-1ex}
\begin{description}
    \item[RQ1] Which topics related to the OneLove armband were discussed on Twitter? (\autoref{ss:topic-modeling}) 
    \item[RQ2] What is the sentiment on Twitter towards FIFA's ban of the OneLove armband? (\autoref{ss:sa}) 
    \item[RQ3] How suitable is a zero-shot or few-shot language model-based pipeline for measuring public sentiment from Twitter/X?  (\autoref{sec:discussion})
\end{description}



\section{Background and Related Work}\label{sec:related-work}

Numerous studies have conducted analyses on Twitter data for FIFA World Cups, including \citet{meier2021politicization}, \citet{patelsentiment}, \citet{nunez2023sentiment},  \citet{hassan2023}, and \citet{fan2020Twitter}. Controversies surrounding countries hosting these events were a focus of previous research. \citet{meier2021politicization} who analyzed Twitter data on the 2018 FIFA World Cup in Russia describe that it was controversial due to the annexation of Crimea in March 2014 and subsequent military conflicts in Ukraine. The authors argue that mega-events are used by host countries to portray a positive national image for domestic and global audiences. They also mention that in the past, countries have come under scrutiny when hosting such events, as civil society has increasingly focused on human rights and civil liberties of host countries, and demanded accountability and political reforms.

\citet{brannagan2018soft} predicted that Qatar intended to use the global attention received from hosting the FIFA World Cup to show Qatar's pursuits of peace, security and integrity, and boister their attractiveness for international tourists. However, the criticism due to the news coverage of alleged human rights violations lead to calls for boycotts and planned protest activities. \citet{hassan2023} found that political discussions surrounding the FIFA World Cup in Qatar initially dominated but then gradually declined, shifting the focus to sporting achievements and cultural exchange. Findings from the sentiment analysis conducted by \citet{nunez2023sentiment} include that more than 66\% of Tweets showed a positive sentiment and that the majority of data contained neutral hashtags, with only around 6\% of Tweets containing a negative hashtags. 

Similarly to \cite{meier2021politicization}, we argue that a sentiment analysis can be an indicator of whether the public is aware of contested issues around mega sports events such as FIFA World Cups and whether protest activities to raise awareness of human rights issues are supported. Further analysis, such as a qualitative content analysis of non-supportive Tweets, can inform future protest activities.  

Differently from previous Twitter analyses on the FIFA World Cup in Qatar, we focus not only on finding out what topics were discussed, but also how users perceived the protest activity of wearing the OneLove armband, a much discussed protest activity. In addition, we analyze German Tweets, while previous research has focused on English Tweets.

\begin{table*}[ht]
\centering
\caption{Performance of different sentiment analysis models. {\normalfont All models were evaluated on our manually labeled test set, as discussed in \autoref{ss:sa}. We link to the publicly available models and indicate if they are multilingual (\img{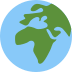}) or only for German ({\img{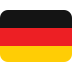}}).}}\label{tab:sent-analysis-performance}
\resizebox{0.8\textwidth}{!}{%
\begin{tabular}{@{}rllrl@{}}
\toprule
&\textbf{Model}                                                              & \textbf{Predicted labels} & \textbf{Acc} {[}\%{]} & $\mathbf{F_1}$ {[}\%{]} \\ \midrule

{\hspace{0.1em}\img{images/earth_africa.png}} & \href{https://huggingface.co/nlptown/bert-base-multilingual-uncased-sentiment}{\tt nlptown/bert-base-multilingual-uncased-sentiment}                                 &   1 to 5 stars            &  46.4    & 19.6    \\
{\hspace{0.1em}\img{images/de.png}} & \href{https://huggingface.co/"oliverguhr/german-sentiment-bert}{\tt oliverguhr/german-sentiment-bert}                                 &   Positive, negative, neutral            &  62.6     & 43.9     \\

{\hspace{0.1em}\img{images/earth_africa.png}} & \href{https://huggingface.co/mistralai/Mistral-7B-Instruct-v0.2}{\tt mistralai/Mistral-7B-Instruct-v0.2}                                 &   In favor, against or neutral            & 64.9      & 47.3     \\
& ~+ 3-shot prompt (examples) &            &  67.1     & 50.7    \\
& ~+ translation to English &            &        69.8      & 54.7     \\
& ~+ Chain of Thought (CoT) reasoning~\citep{kojima2205large} &            & 74.3& 61.5    \\
& ~+ all three techniques &            & {76.1}      & {64.2}     \\
{\hspace{0.1em}\img{images/earth_africa.png}} & {\tt gpt-3.5-turbo}\footnotemark[1]  (3-shot prompt, translation, CoT)                               &   In favor, against or neutral            & 59.9     & 39.9      \\
{\hspace{0.1em}\img{images/earth_africa.png}} & {\tt gpt-4-turbo}\footnotemark[1]  (3-shot prompt, translation, CoT)                               &   In favor, against or neutral            & \textbf{80.2}     & \textbf{70.3}      \\

\bottomrule
\end{tabular}
}
\end{table*}

\section{Method}
All analyzed Tweets were collected by the authors through the formerly freely available, official Twitter API. This API allowed access to all public Tweets of the last 30 days. To find Tweets with certain characteristics, we used the following filter: 

{\small
\begin{verbatim}
-is:retweet -is:reply -is:quote lang:de 
-liveticker -newsticker (#WM2022 OR 
#FIFAWorldCup OR FIFA OR WM OR (WM (Katar 
OR Qatar)) OR ((Fußball OR Fussball)(Qatar 
OR Katar OR Weltmeisterschaft)))
\end{verbatim}
}

This string contains the German and English spelling variant of Qatar and Football. Weltmeisterschaft (abbr. WM) is the translation for "World Cup" in German. 
These hashtags and words were selected to ensure that Tweet topics are connected to the World Cup. Tweets that only contain content about football or only Qatar should be excluded through this search. Retweets, replies and quotes were excluded from the collection of data, as context needs to be taken into account when interpreting such Tweets (see Limitations). The language was set to German and liveticker and newsticker were excluded to avoid Tweets only about news regarding this event, as they are not relevant for sentiment analysis. The data set consists of 132,150 Tweets that were Tweeted in a time period from November 20 to December 18, 2022. 

\subsection{Topic modeling}\label{ss:topic-modeling}
To get an overview of important topics in our data set, we performed unsupervised topic modeling. 
We used BERTopic~\citep{grootendorst2022bertopic} with a multilingual model 
and a CountTokenizer with n-grams of length 1 to 3, to also include bi-grams like `one love' and tri-grams like `One Love Binde'.
The BERTopic library only includes stopwords in English, so we iteratively wrote our own short list to improve the clusters 
(see \autoref{appendix:stopwords}).

We validated the discovered topics using a separate test set, where we manually annotated 600 Tweets belonging to seven topics: irrelevant, game Tweet, news, boycott, human rights, OneLove Binde and general politics. 
%
%
The highest amount of Tweets were news and reporting of the game progress, which are not interesting for our analysis. 

\begin{figure}[t]
    \centering
    \includegraphics[width=\linewidth]{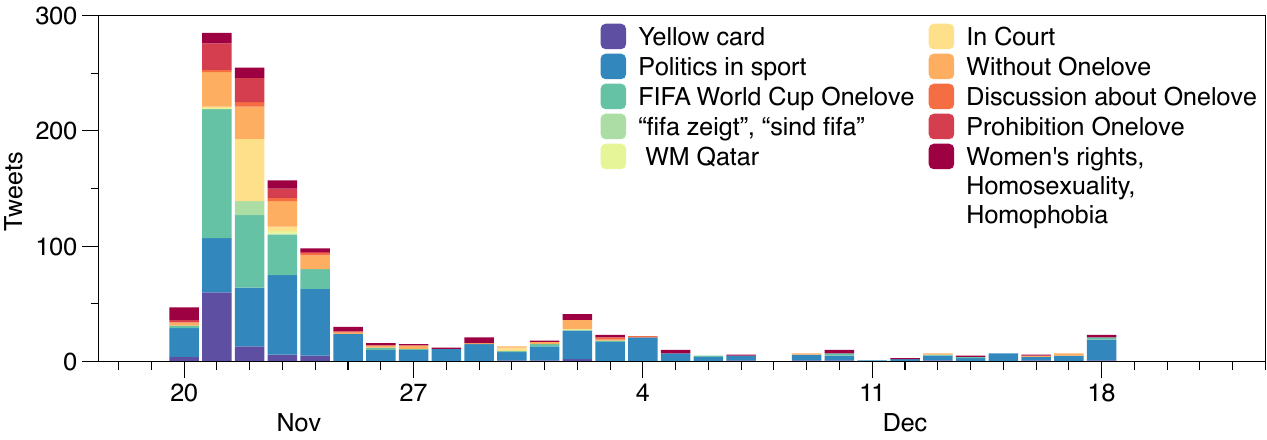}
    \caption{Timeline of discussed topics related to OneLove. {\normalfont The topics are found by topic modeling (\autoref{ss:topic-modeling}) and linked to OneLove with manual annotation of a subset of Tweets.}}
    \label{fig:subfig1}
\end{figure}

The OneLove armband was relevant to Twitter users and our topic model found this topic and the issues related to this. As mentioned before, due to the media attention and controversy surrounding this protest activity, we decided to focus on Tweets about the OneLove armband.  We manually annotated 200 Tweets that contain the word OneLove or a spelling variation (one-love and one love) as being for, against or neutral towards wearing the armband. Another of the authors annotated 100 Tweets to calculate Inter-Annotator Agreement (IAA). Of the 200 annotated Tweets on OneLove, 88 were annotated as for OneLove and 20 against OneLove. 
We did this by selecting the topics from our BERTopic pipeline that were associated with the Tweets that we manually labeled as `OneLove Binde' (see \autoref{fig:subfig1}).

The topics found by our topic modelling pipeline align well with different aspects related to the OneLove armband, such as the penalty for wearing the armband despite the ban (a yellow card), discussions about politics in sports and various OneLove-related topics.

\footnotetext[1]{GPT-3.5 and GPT-4 are not fully supported with {\tt select}-queries in Guidance, so unconstrained generation is used and this affects the accuracy.} 




\subsection{Few-shot sentiment analysis}\label{ss:sa}
We used sentiment analysis to analyze the Tweets on the OneLove topic, which is a commonly used method to extract subjective information, such as opinions and attitudes~\citep{medhat2014sentiment,mantyla2018evolution}. 
%
%
However, our task is to gauge support for wearing the OneLove armband, which is distinct---and often the opposite---of support for the ban that FIFA implemented.
This task, often referred to as stance detection~\citep{alturayeif2023systematic}, is different than what publicly available sentiment analysis models are trained on, so we perform an evaluation of different language models.
In addition, we focus on evaluating zero-shot and few-shot methods to answer RQ3. Because of this focus, our approach could be directly applicable for new studies following other events, which in turn also allows for a faster reaction to develop a pipeline while an event is still unfolding.

We first create a set of gold labels by having two native German speakers annotate 148 randomly sampled Tweets using Doccano\footnote{\url{https://doccano.github.io/doccano/}}.
Our annotation labels are `in favor of the OneLove Binde' ($n=63$), `against the OneLove Binde' ($n=15$) and  `neutral' ($n=70$).
Some of the Tweets did not directly contain an opinion in the text, but used tags like \#WMderSchande (World Cup of Shame) and \#BoycottQatar. Other Tweets were commenting on how the OneLove armband is only a replacement for a real rainbow-colored armband and a compromise, these Tweets were also annotated as being for the armband.
The inter-annotator agreement is $\kappa=0.68$, which indicates significant agreement~\citep{viera2005understanding}.

We evaluate two already finetuned BERT-based models and validate them by calculating macro-averaged accuracy and micro-averaged $F_1$ scores between the predicted labels and the gold labels.
However, both models are finetuned for sentiment analysis on different domains, mostly reviews with a star-based rating system. As a consequence, the observed error rates are too high when applied to our dataset, as shown by the $F_1$ scores of 20\% and 44\% in \autoref{tab:sent-analysis-performance}. 

To address this, we test a similar setting using generative LMs, where we tested Mistral, GPT-3.5 and GPT-4. 
For the publicly available model Mistral 7B~\citep{jiang2023mistral}, we evaluated several variations of a prompt: (i) only the base prompt with the coding instructions, (ii) with 3 examples for 3-shot classification, (iii) with additional instructions to translate the Tweet to English, (iv) Chain of Thought (CoT, ``\emph{let's think step by step}'') reasoning~\citep{kojima2205large} and (v) a combination of all prompts. We also test the conversation-tuned GPT-3.5 and GPT-4 models by OpenAI using all aforementioned prompting techniques, as they proved to increase the performance on the Mistral model. 
We also note that the models struggle most with the `against OneLove binde' class, for instance GPT-4 has an AUC score of 0.697 for this class, while the others are slightly higher with 0.769 (in favor) and 0.747 (neutral).  

Based on this evaluation (see \autoref{tab:sent-analysis-performance}), we find that an optimized prompt with examples, Chain of Thought reasoning and a translation all improve the Mistral-based labeling. Interestingly enough, a BERT-based classifier performs almost as well (62.6\% accuracy) as Mistral without prompting techniques (64.9\%), at a lower inference cost. Nevertheless, by optimizing the prompt Mistral achieves 76.1\% accuracy, which outperforms GPT-3.5 and is only slightly worse than the relatively expensive GPT-4 model, so we use Mistral with all prompting techniques to classify all Tweets on OneLove.


\section{Discussion}\label{sec:discussion}

\paragraph{Opinions on the OneLove armband.} We first analyzed the topics found in the Tweets, where we found multiple topics related to OneLove (see \autoref{fig:subfig1}).
There are more Tweets about the game itself, but `one love' and `boycott' are the first political topics we found with our topic modeling.
The range of topics is broad, and include for instance discussions of the consequences of wearing the OneLove armband, women's rights and politics in sports.
Initially, the discussion was about OneLove and related topics, such as women's rights, homosexuality and homophobia, as well as the mention of `legal action'.
Interestingly enough, most of the Tweets after the ban, on November 21, 2022, shifted to focus on politics in sports in general, as opposed to the armband itself or banning it. 

\begin{figure}
    \centering
        \includegraphics[width=\linewidth]{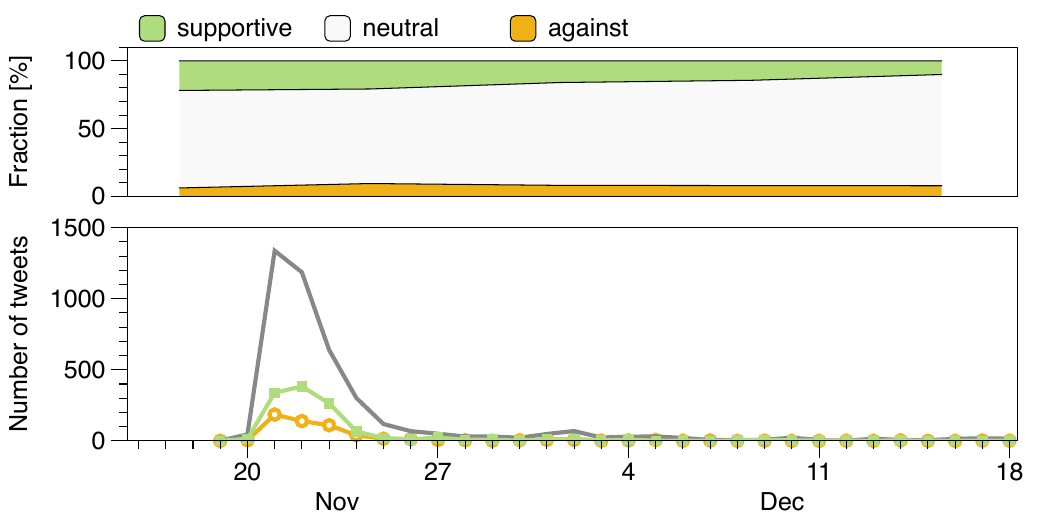}
        \caption{Sentiment of Tweets related to OneLove.}
        \label{fig:subfig2}
    \end{figure}

The sentiment of the Tweets slightly shifted towards a more neutral stance, however there are more supportive Tweets for wearing the OneLove armband (see \autoref{fig:subfig2}).
We also observe an initial spike of Tweets as a reaction on the ban of the OneLove armband, but this quickly ebbed away.
The overall sentiment correlates with some surveys, e.g. \citet{morgenpost_umfrage}, although we observe more Tweets taking a neutral stance.

\paragraph{Few-shot analysis pipeline.}
In this paper, we also presented a few-shot pipeline to cluster Tweets using topic modeling and analyze them using zero-shot and few-shot language models. 
We found that both aspects of the pipeline work sufficiently well to use for events and crises: First, the topics discovered by BERTopic aligned with our manual labels (see \autoref{ss:topic-modeling}). Second, large language models, such as Mistral-7B, are surpassing finetuned BERT models (see \autoref{ss:sa}) and allow for in-context learning with few examples~\citep{kojima2205large}, meaning classification pipelines are quicker to set up.

\paragraph{Comparison to Surveys}

Some social media users do not post their opinion publicly or post nothing, even if they have an opinion. The imagined audience, lurking behavior and deletion of Tweets impact the availability of social media data \cite{litt2016imagined, gong2016unravelling, almuhimedi2013_deleted, zhou2016tweet}. Our analysis can be used as an additional data point alongside other sources of information. Many surveys conducted by newspapers and other media outlets observed similar trends in the fraction of people supporting a boycott as our analysis does. A 2022 survey conducted by a local newspaper shows that 46\% of German citizens who responded intended to boycott the 2022 FIFA World Cup, while 28\% were against a boycott \cite{morgenpost_umfrage}. Another survey's results showed that 72\% of respondents were in favor of a boycott \cite{ndr_umfrage}. There was also a non-representative survey which German government officials responded to and approximately half favored a boycott \cite{merkur_umfrage}. 

While surveys with fixed responses can give an overview of agreement and disagreement with an issue, an advantage of social media data compared to such surveys is that it is possible to analyze how individuals form or explain their stance on a topic. Topic modeling and sentiment analysis can also be combined with a qualitative analysis of Tweets if particular topics are of interest. Further analyses could focus on whether there are any group differences among those supporting or boycotting protest activities, as previous research has focused on inferring demographic data from information provided on Twitter \cite{sloan_inferring_social2013, Culotta_Kumar_Cutler_2015, sloan2015tweets}.   

\section{Conclusion}
The FIFA World Cup 2022 in Qatar was accompanied by controversies, among them allegations of human rights violations of migrant workers and LGBT people. Football teams planned to wear the OneLove armband as a protest activity, however the FIFA threatened to sanction wearing the armband. 
We conducted topic modeling and found that Twitter users talked about the ban of the OneLove armband for a few days after the ban. We then focused on Tweets about OneLove and the subsequent ban of the OneLove armband, to gauge whether Tweets were for or against wearing the armband. Our analysis shows that there was more support for wearing the armband than not, although the support did fade over time, which could indicate that Twitter users perceived the protest as unsuccessful. 
We identified a shift from the armband-specific discourse to a broader discussion on politics in sports. 
A purpose of this protest activity was to raise awareness of human rights violations. 
If a sentiment analysis reveals negative public perception towards a particular protest activity, this suggests a potential need for reassessment and modification of the protest strategy which is possible while an event takes place through our proposed few-shot pipeline.

\section*{Limitations}

The data set we analyzed consisted of German Tweets, which means that results only allow us a glimpse into German-speaking Twitter users' opinions on the FIFA World Cup in Qatar. X, formerly known as Twitter, has users from certain demographics and is not used by every German-speaking person, thus the results are not generalizable to all German-speakers. We did not try to infer demographics from our data. Nonetheless, our work contributes to research on social media data centered on languages other than English and goes beyond Germany's borders, as German is also spoken in other countries, such as Austria and Switzerland.


We elected to not include replies, which means there might be posts in threads that are for or against the OneLove armband that we did not cover with our analysis. Since the added context (the original Tweet) might make interpreting these Tweets more difficult\footnote{Is positive or negative sentiment towards the ban or the OneLove Binde or towards the original Tweet?}, we decided to leave out these replies. Future work could focus to include these replies with the original Tweet.

Finally, we evaluated multiple models to classify Tweets. \citet{bruyne2021} showed that different BERT-based models performed differently with regard to emotion detection or sentiment analysis, which might be the case for the models we tested as well. Biases in the model might affect the classification as well~\citep{waseem2021disembodied}. However, an analysis of these issues for the models we used is out of scope for this work.

\section*{Ethical considerations}
We only used publicly available data and did not interact with human subjects, which means our work did not classify as human subjects research by our IRBs. Since this research did not seek informed consent, data were taken from a public online venue. On X, formerly known as Twitter, users have different privacy options, such as posting public and private Tweets. We collected only public Tweets which were easily accessible for all users through the search function. Following the conditions of the Twitter API, we cannot publish the full dataset that we used, instead we publish a ``dehydrated'' version 
which requires access to the API to receive full information on each Tweet. For our analysis and collection of data, we only used the Tweet, no further information on the person who posted it was collected.

\section*{Acknowledgments}
%
Pieter Delobelle received funding from the Flemish Government under the ``Onderzoeksprogramma Artificiële Intelligentie (AI) Vlaanderen'' programme and was supported by the Research Foundation - Flanders (FWO) under EOS No. 30992574 (VeriLearn). He also received a grant from ``Interne Fondsen KU Leuven/Internal Funds KU Leuven''.

\bibliography{slr_references.bib}

\newpage
\appendix

\section{Stopwords}\label{appendix:stopwords}
\begin{itemize}
  \setlength\itemsep{-0.3ex}
\item der
\item dem
\item das
\item und
\item er
\item ein
\item den
\item ist
\item es
\item die
\item ich
\item man
\item zu
\item nicht
\item so
\item wie
\item was
\item auch
\item aber
\item wenn
\item als
\item noch
\item mal
\item sich
\item dass
\item nur
\item oder
\item dann
\end{itemize}

\onecolumn
\section{Prompts}
\begin{lstlisting}[caption=\textbf{Prompt for Mistral-7B}]
[INST] Rate German Tweets as 'supportive', 'against' or 'neutral' of wearing the One Love bracelet or one love binde in German. These Tweets were posted during the world cup in Qatar.

Coding instructions:
- Critique for the FIFA policy of banning the bracelet or wishing players could wear it should be interpreted as support for the one love binde. 
- Similarly, if support for the one love binde is slightly implied, that is sufficient for it to be supported.
- If the tweet is factual without any sentiment, consider it neutral.
- If the tweet uses negative connotations with the one love binde and what it stands for (the LGBT community), the tweet is against.

Tweet: "lul und ich werde angefeindet weil ich den tv anmache hugo lloris weigert sich bei der anstehenden fußball weltmeisterschaft in katar die one love binde zu tragen er wolle die kultur des gastgebers respektieren so die begründung des frankreich kapitäns boykottqatar" [/INST]
Support for wearing the one love binde: supportive
[INST] Tweet: "dfb torwart neuer hält an one love binde bei wm in katar fest" [/INST]
Support for wearing the one love binde: neutral
[INST] Tweet: "soso die scheiß will wohl die onelove binde verbieten ihr habt echt nicht mehr alle latten am zaun infantino und seine geldsäcken sollte man in die wüste schicken fifaworldcup qatarworldcup qatar dopa" [/INST]
Support for wearing the one love binde: against
[INST] Tweet: "{tweet}" [/INST]
English translation:  {translation}
[INST] Now look at the previous tweets and analyze the similarities with the latest tweet that we want to label and if that tweet expresses support for wearing the one love binde. Let's think step by step:[/INST] {reason}
[INST] Therefore the tweet expressed the following sentiment towards wearing the one love binde:[\INST] {decision} 

\end{lstlisting}

\end{document}